\documentclass[conference]{IEEEtran}
\usepackage{cite}
\usepackage{amsmath,amssymb,amsfonts}
\usepackage{enumitem}
\usepackage{graphicx}
\usepackage{textcomp}
\usepackage{algorithm}
\usepackage{algpseudocode}
\usepackage{xcolor}
\def\BibTeX{{\rm B\kern-.05em{\sc i\kern-.025em b}\kern-.08em
    T\kern-.1667em\lower.7ex\hbox{E}\kern-.125emX}}
\begin{document}
\title{Blockchain-enabled Clustered and Scalable Federated Learning (BCS-FL) Framework in UAV Networks}
 \author{\IEEEauthorblockN{Sana Hafeez, Lina Mohjazi, Muhammad Ali Imran and Yao Sun\\}
\IEEEauthorblockA{\textit{James Watt School of Engineering, University of Glasgow, Glasgow, United Kingdom} \\
     Email: {\{s.hafeez.1\}@research.gla.ac.uk},\\ {\{Lina.Mohjazi,   Muhammad.Imran, Yao.Sun\}@glasgow.ac.uk}}} 
\maketitle
\begin{abstract}
Privacy, scalability, and reliability are significant challenges in unmanned aerial vehicle (UAV) networks as distributed systems, especially when employing machine learning (ML) technologies with substantial data exchange. Recently, the application of federated learning (FL) to UAV networks has improved collaboration, privacy, resilience, and adaptability, making it a promising framework for UAV applications. However, implementing FL for UAV networks introduces drawbacks such as communication overhead, synchronization issues, scalability limitations, and resource constraints. To address these challenges, this paper presents the Blockchain-enabled Clustered and Scalable Federated Learning (BCS-FL) framework for UAV networks. This improves the decentralization, coordination, scalability, and efficiency of FL in large-scale UAV networks.
The framework partitions UAV networks into separate clusters, coordinated by cluster head UAVs (CHs), to establish a connected graph. Clustering enables efficient coordination of updates to the ML model. Additionally, hybrid inter-cluster and intra-cluster model aggregation schemes generate the global model after each training round, improving collaboration and knowledge sharing among clusters. 
The numerical findings illustrate the achievement of convergence while also emphasizing the trade-offs between the effectiveness of training and communication efficiency.
\end{abstract}


\section{Introduction}
Federated learning (FL) has emerged as a privacy-preserving approach for collaborative machine learning (ML) without direct data sharing \cite{unknown}. In FL, models are locally trained on individual devices using their respective data, with only model updates aggregated to enhance a shared global model. This enables collaborative learning while safeguarding data privacy \cite{10182294}.

Coordinating a large, decentralized network of heterogeneous unmanned aerial vehicles (UAVs) for FL poses a key challenge. Existing approaches often rely on a centralized server to orchestrate participant roles, aggregate model updates, and allocate rewards. However, this introduces vulnerabilities, trust issues, and inaccuracies in reward allocation. Recent studies have explored decentralized clustering schemes to group UAVs and rotate cluster heads (CH) to reduce communication costs \cite{Mohsan2023}. Nonetheless, these methods still depend on centralized components for scheduling and global model aggregation.

Addressing this limitation, Al et al. introduce a decentralized FL framework that uses merged UAV clusters to improve energy efficiency \cite{al2023decentralized}. However, their framework retains the need for centralized scheduling in model aggregation. Alternatively, in \cite{qu2021efficient}, the authors formulate the efficient edge intelligence clustering problem for UAV swarms (e-EIC) and propose an iterative algorithm employing optimal policy and local search techniques.

To improve energy efficiency, Qu et al. \cite{qu2021decentralized} propose a clustering scheme for UAVs in FL. Unfortunately, this approach relies on a leading UAV for coordination, creating a single point of failure (SPoF) vulnerability. Recent advancements in blockchain technology offer promise in mitigating these challenges \cite{hafeez2022beta}. Using blockchain, decentralized incentive mechanisms through smart contracts can motivate UAVs to contribute resources for FL, thereby enhancing security, transparency, and credibility.
Exploring the scalability of FL to large UAV swarms, Hou et al. \cite{hou2023uav} examine the challenges posed by central server reliance for model aggregation, especially for massive remote UAV networks with limited resources.

Bridging the introductory motivation and our technical framework, we first summarize the key components of our proposed model. The UAV network is organized into clusters based on proximity. Each cluster has a head UAV to facilitate aggregation and coordination. The CH forms an interconnected network to enable model updates between clusters. Smart contracts handle registration, cluster formation, and decentralized aggregation. The core ideas involve UAV clustering, CH hierarchy, inter-cluster relationships, blockchain integration, and FL workflow. With this high-level prelude, we now present the specifics of our model tailored for the FL UAV swarm.

In light of identified limitations in current FL implementations, we introduce a comprehensive hybrid clustering method for diverse UAV swarm topologies, even accommodating UAVs positioned beyond the communication range. In this method, $\mathrm{CH}$ UAVs establish a connected graph, ensuring interconnectivity among clusters. We present two model aggregation schemes within a clustered network. Fully Centralised Aggregation (FCA) and $k$-hop Aggregation ($k$HA). Numerical evaluations demonstrate the convergence achieved and the reduction in communication overhead across the network.
The remaining parts are organized as follows:
Section II describes our system model for UAV networks.
Section III outlines the blockchain-enabled FL framework, which includes dual model aggregation approaches.
In Section IV, we present the results of our numerical simulations and provide a detailed discussion.
Finally, we conclude the paper in Section V.
\section{System Model for UAV Networks }
In this section, we begin by introducing the UAV network topology mode and providing a comprehensive overview of the framework's key components. Subsequently, we delve into the details of our proposed clustering scheme, specifically tailored for UAV swarm networks.
\subsection{UAV Network Topology Model}
In our study, we focus on a UAV swarm comprising $U$ UAVs deployed within a designated geographic region. This swarm collectively engages in ML model training using FL techniques. Individual UAVs are denoted by the set $\mathcal{U} = {1, \ldots, U}$. Each UAV $u$ maintains a constant altitude and is characterized by coordinate in a two-dimensional space. The position of the UAV $u$ is represented as $\mathbf{p}_u = (i_u, j_u)$, where $\mathbf{p}_u$ signifies the position of the UAV $u$. Throughout the FL process, the UAVs strive to remain within the locations defined by $\mathbf{p}_u$.

However, certain unforeseen circumstances, such as adverse weather conditions, can lead to a UAV deviating from its intended location $\mathbf{p}_u^d$ during the $d$-th training round. This deviation is measured as $\delta = |\mathbf{p}_u^{d+1} - \mathbf{p}u^d|$, where $\delta$ represents the assumed distance measured at round $d$ using the Euclidean norm $||\cdot||$. To ensure collision avoidance and maintain a smooth flight trajectory, we introduce a maximum allowable distance denoted as $\delta{\text{max}}$. This limit restricts the extent to which UAV $u$ can diverge from its initial position $\mathbf{p}_u^0$.
  \begin{figure*}[htb!]
\centering    \includegraphics[width=0.64
\textwidth]{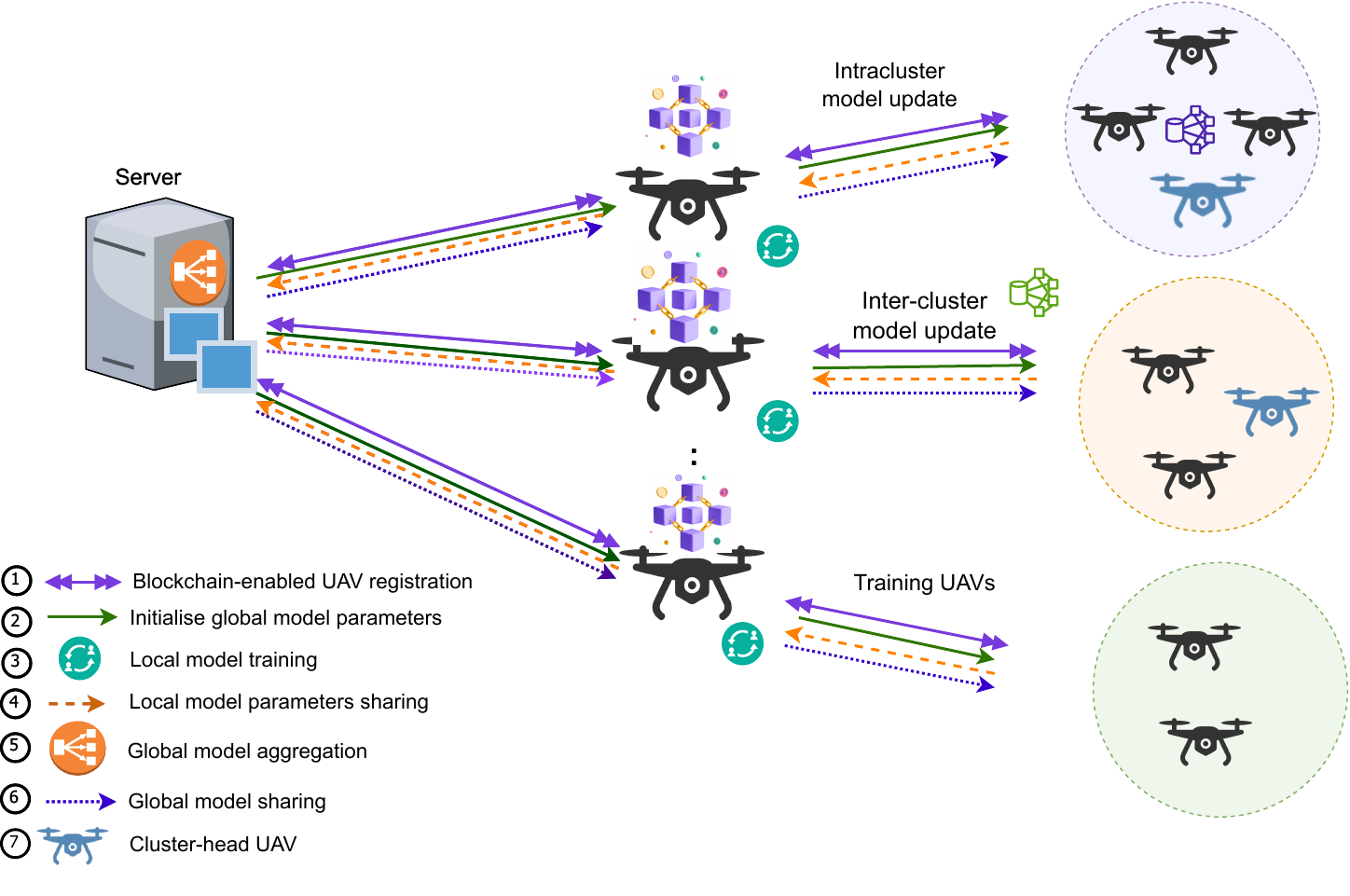}
\caption{The BCS-FL Framework.}
\label{fig}
\end{figure*}
In addition, we make an assumption regarding the communication capabilities of each UAV within the swarm. Specifically, we assume that every UAV is equipped with a maximum communication range denoted as $R_{\max}^{\text{com}}$. This communication bandwidth allows UAVs to exchange data with nearby counterparts, facilitating regional model updates during FL. We make a further assumption that the UAVs within the swarm are sufficiently dense, ensuring comprehensive coverage of the sensing area and interconnectivity between the UAVs while adhering to the constraints defined by $R_{\max}^{\text{com}}$ and $\delta_{\max}$. This notion enables each UAV in the network to potentially establish communication links with any other UAV, utilizing either direct connections or multiple intermediate connections.
To streamline the clustering process, UAVs utilize beacon frames to discover and identify other UAVs in close proximity, thus forming distinct clusters \cite{abdulhae2022cluster}. The aggregation of these clusters is denoted as the set $\mathcal{Q} = {1, \ldots, q, \ldots, Q}$. In our subsequent discussions, we denote a UAV node by $u_q$ if it pertains to a node within cluster $q \in \mathcal{Q}$. Furthermore, we define the set of UAVs within cluster $q$ as $\mathcal{U}_q$. Moreover, we assume that each cluster $q$ includes a designated $\mathrm{CH}$ UAV, identified as $u_q^{\text{CH}}$, entrusted with coordination of communication between clusters.

In real-world UAV deployments, it's essential to acknowledge that direct communication links might not be established between all UAVs. This scenario could arise due to factors such as limited communication range, physical obstructions, or the decentralized nature of the UAV network. Consequently, complete collaboration across all UAVs might only be viable within each connected subgraph of the network. In our study, we focus on scenarios where the blockchain-enabled clustered and scalable federated learning framework (BCS-FL) can be easily applied to each connected subgraph, omitting the complexities associated with unlinked UAVs.

Our objective is to explore the potential of our framework and evaluate its effectiveness in facilitating seamless communication and collaboration among UAVs operating within these interconnected subgraphs.
\subsection{Clustering Architecture for UAV Networks}
To facilitate efficient communication in FL across extensive UAV networks, our focus is on optimizing the aggregation of local model updates. Specifically, we introduce a hierarchical clustering strategy aimed at managing communication overhead during model aggregation within each training round.

Our primary objective is to determine the optimal number of UAV clusters, denoted as $Q$. This optimal value $Q$ serves the dual purpose of minimizing inter-cluster communication overhead while upholding connectivity between CHs. Our approach employs an iterative clustering algorithm that progressively increases $Q$ through the utilization of $k$-means clustering, with an initial value $Q$ as the starting point.

The algorithm ends when CH connectivity is achieved with the minimum value $Q$. To account for potential UAV movements during training and ensure the connectivity of CH UAVs, we incorporate the distance parameter $\sigma = R_{\text{max}}^{\text{com}} - 2\delta_{\text{max}}$. By hierarchically clustering the extensive UAV network and optimizing the number of interconnected CH UAVs, our proposed approach significantly enhances the communication efficiency of model aggregation within FL.

\section {Blockchain-based Federated Learning}
BCS-FL, is a novel framework that offers advantages for UAV networks, primarily preserving data privacy while enabling collaboration, This is accomplished through secure data storage and communication protocols, ensuring sensitive information remains protected during FL. In the following, we outline the key components and mechanisms, including two model aggregation strategies FCA and $k$HA.
\subsection{ BCS-FL Overview}
Our proposed framework, depicted in Fig. \ref{fig}, introduces BCS-FL in UAV networks. The BCS-FL framework utilizes smart contracts as automated modules on blockchain. These contracts play a vital role in cluster formation, UAV registration, and model aggregation. The utilization of smart contracts enables UAV registration and cluster creation based on proximity. Authorized users, such as UAVs for registration and training, manage these processes.

During the model aggregation phase, smart contracts compute weighted averages of local model updates submitted by training UAVs. This aggregation results in the creation of an aggregated global model, which is subsequently distributed to the participating training UAVs. The framework involves three distinct groups of UAVs: blockchain-assisted UAVs, registration UAVs, and CH UAVs. Each group has a critical role in the collaborative training process.
CH UAVs engage in intercluster communication to obtain a global model by exchanging their locally aggregated models. At the start of each training iteration, the global model is transmitted to the training UAVs, ensuring continuous refinement of the model throughout the process.

The significance of registration UAVs becomes evident in their role of orchestrating cluster formation across the network landscape. These UAVs utilize beacon frames to identify and discover neighboring UAVs, ultimately facilitating the creation of distinct clusters. Through this clustering mechanism, registration UAVs contribute to the efficient organization and communication within the framework. The terms discussed in the provided text relate to a contextual framework that involves a blockchain-focused architecture designed for the execution of FL in collaboration with UAVs. Here, we provide an in-depth explanation of these terms:
\begin{enumerate}
\item Blockchain Registered UAVs:
In the smart contract in Algo. 1, defines a UAVNode struct to represent participating UAV nodes and their attributes. The contract stores UAV nodes in a mapping, retrievable by ID. The joinBCSFL function allows authorized users to register new UAV nodes on the blockchain network. The contract uses a mapping to store and retrieve UAV nodes based on their ID. The join BCS-FL function allows an authorized user to register a new UAV node in the BCS-FL network, and the getNodeById function enables retrieving node information by its ID.
\begin{algorithm}
\caption{BCSFLContract}
\begin{algorithmic}[1]
\State \textbf{struct} UAVNode:
\State \hspace{0.5cm} address owner
\State \hspace{0.5cm} string nodeId
\State \textbf{mapping} (uint256 $\rightarrow$ UAVNode) uavNodes
\State uint256 totalNodes
\State \textbf{event} NodeJoined:
\State \hspace{0.5cm} uint256 id
\State \hspace{0.5cm} address owner
\State \hspace{0.5cm} string nodeId
\State \textbf{function} joinBCSFL(string nodeId):
\State \hspace{0.5cm} uint256 newNodeId $\gets$ totalNodes
\State \hspace{0.5cm} UAVNode newNode $\gets$ uavNodes[newNodeId]
\State \hspace{0.5cm} newNode.owner $\gets$ msg.sender
\State \hspace{0.5cm} newNode.nodeId $\gets$ nodeId
\State \hspace{0.5cm} totalNodes++
\State \hspace{0.5cm} \textbf{emit} NodeJoined(newNodeId, msg.sender, nodeId)
\State \textbf{function} getNodeById(uint256 id):
\State \hspace{0.5cm} UAVNode node $\gets$ uavNodes[id]
\State \hspace{0.5cm} \textbf{return} node.owner, node.nodeId
\end{algorithmic}
\end{algorithm}
\item {Configuration of UAV}:
We emphasize the typical FL task of using numerous drones for training, denoted by $U$. Local datasets $\mathcal{R}_u$  consisting of  $\left|\mathcal{R}_u\right|$ data samples are accessible to each training UAV $u$. For a given model parameter vector ${f}_l$, the loss is quantified by the function ${l}$ and a loss sample $\zeta$ has its local objective function $O_u({l})$ for training UAV $u$ is defined as
\begin{equation}
    O_u({l})=\frac{1}{\left|\mathcal{R}_u\right|} \sum_{\zeta \in \mathcal{R}_{l}} O({l}, \zeta). 
\end{equation}
The global objective function $O{(l)}$ with respect to $U$. The formal statement of the training clients can be expressed as follows
\begin{equation}
O_{\mathbf{l}}=\sum_{u=1}^U\chi_u O_u \mathbf{l},
\end{equation}

In a training round $d$, a UAV $u$ belonging to cluster $q$ conducts stochastic gradient descent (SGD) updates on its dataset $\mathcal{R}u$, resulting in an updated weight vector ${l}_{u, c}^t$. In the following sections, we will look into the importance of $\mathrm{CH}$ UAVs in aggregating local model updates and elucidate the mechanism through which they contribute to the ultimate global model. The weight $\chi_u$ corresponds to the training UAV $u$.
\item {Cluster-head UAV Swarms}:
  $\mathrm{CH}$ UAVs have distinct roles within each cluster, differing from the training UAVs. They do not participate in local model training but are responsible for the following tasks:
\begin{itemize}
  \item Receiving local model updates: The 
  $\mathrm{CH}$ UAV receives updates of local models from the training UAVs within its cluster. Moreover, at the onset of each training round, it takes the responsibility of distributing the newly aggregated global model to all the participating training UAVs.
  \item Enabling intercluster model aggregation: The 
  $\mathrm{CH}$ UAV facilitates the aggregation of models between clusters by interacting with 
  $\mathrm{CH}$ UAVs from other clusters. UAVs establish a network where each $\mathrm{CH}$ UAV is linked to another, ensuring seamless communication and coordination among them.
\end{itemize}
Ensuring effective communication demands high node centrality for the $\mathrm{CH}$ UAVs within their respective clusters. Placing them in proximity to the cluster center is crucial to maximize their impact. In the context of distributed ML, the relationships between node centrality, communication efficiency, and the federated averaging (FedAvg) aggregation algorithm \cite{10.1145/3432291.3432303} are interconnected. CH UAVs exhibit higher centrality, which in turn bolsters communication effectiveness. FedAvg capitalizes on this hierarchy by prioritizing influential nodes during aggregation, thereby fostering efficient collaboration and convergence toward the final model.

The synergy among node centrality, effective communication, and the FedAvg aggregation algorithm lies in their collective contributions to enhancing the efficiency of collaborative training within UAV networks.

Here, $l_q^d$ represents an aggregated model of cluster $q$.
\end{enumerate}
\begin{equation}
l_q^d = \sum_{u \in \mathcal{U}q, u \neq u_q^{\mathrm{CH}}} \chi_u {l}{_u^d,q},
\end{equation}

where $\chi_u$=$\frac{|\mathcal{R}_u|}{|\mathcal{R}_q|}$ is the ratio between the data samples on UAV $u$ and the data samples of the entire cluster $q$, adhering to the FedAvg.
\subsection{Strategies for Intercluster Aggregation}
During ML model training, effective utilization of available data is crucial. In our decentralized FL framework, we enable the global model to efficiently incorporate learning from local models across various clusters. This fosters rapid convergence and improved training performance. Nevertheless, this approach might lead to a notable increase in message exchanges among CH UAVs, resulting in substantial communication overhead. To address this challenge, we introduce two aggregation strategies, each offering distinct trade-offs between training performance and overhead. These strategies are known as FCA and $k$HA.
\begin{itemize}
   \item  Fully Centralized Aggregation (FCA): This strategy forms its basis on incorporating model updates from every CH UAV during each training round. Formally, the global model representing the entire network is defined as
\begin{equation}
\check{l}^d = \frac{1}{Q} \sum_{q \in \mathcal{Q}} l_q^d.
\end{equation}
In practice, in each training round, one CH UAV is randomly selected to receive model updates from all other CH UAVs. This selected CH UAV then calculates the aggregated global model as described in Equ. 4.
   \item  $k$-Hop Aggregation ($k$HA):
 To address the issue of communication overhead caused by the widespread distribution of UAVs over a vast geographical area during model aggregation, we propose the $k$HA strategy. This approach permits each CH UAV to share its locally aggregated model with neighboring CH UAVs located within a maximum distance of $k$ hops. Refer to Fig. \ref{kHA} for illustration, where each hop is represented by $k$. For example, when $k$ is set to 2, the source CH UAV in red transmits its locally aggregated model to seven neighboring CH UAVs within a maximum of two hops (depicted by the purple, orange, and green CH UAVs in Fig. \ref{kHA}).
\begin{enumerate}
      \item The Initial setup: Training UAVs retrieve the globally aggregated model from the previous round. At the start of each training iteration, participating UAVs receive the initial dataset from their respective CH UAVs, forming the basis for the training process.
     \item Formation: Instruction UAVs independently update their local models using the SGD method and their own local datasets.

     \item Intracluster Aggregation: Training UAVs within each cluster communicate their relevant CH UAV to their respective local model vectors. CH UAVs then use local model aggregation to derive an aggregated model representation for the cluster.
   \item Intercluster Aggregation: CH UAVs exchange models among themselves, sharing locally aggregated models based on the selected strategy (FCA or $k$HA). This enables the computation of the globally aggregated model by integrating models across different clusters, preparing all UAVs for the next training phase.
\item {Repeat:} The process of local model training on each UAV, followed by aggregated global model updates, is iterated until the training convergence criteria are met or the maximum predefined number of global rounds is reached.

By employing this workflow, our FL framework promotes effective collaboration and iterative model refinement among UAVs, ultimately driving the learning algorithm toward convergence or the completion of the maximum number of training rounds.
\end{enumerate} 
\begin{figure}[htbp]
\centering
        \includegraphics[width=0.30\textwidth]{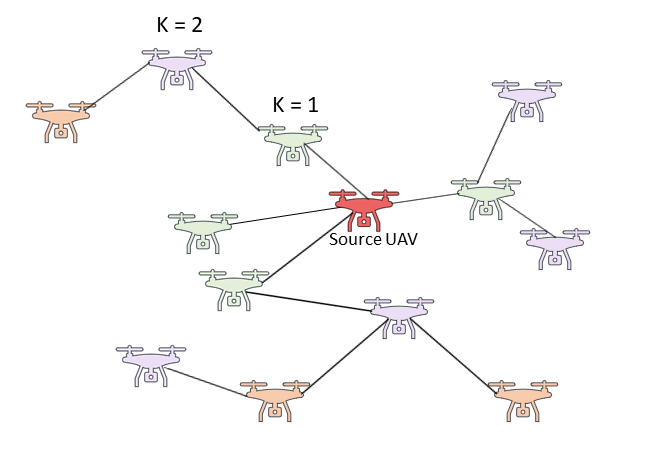}
\caption{A Demonstration of the kHA Scheme.}
\label{kHA}
\end{figure}
\end{itemize}
\section{Numerical Results and Discussions }
\subsection{Simulation Settings }
We deploy UAVs at random locations within a 1000 m × 1000 m area, ensuring they form a connected graph based on predefined $R_{\max}^{\text{com}}$ and $\delta_{\text{max}}$ values. We set $R_{\max}^{\text{com}}$ to 150 $m$ and $\delta_{\max }=5 m$ in our simulations. All training UAVs are actively involved in each global round. We evaluate using two commonly benchmark datasets, namely MNIST and CIFAR-10 \cite{hoang2023clustered},  to assess the effectiveness of our proposed schemes. Using two dataset partitioning strategies, independent and identically distributed (IID) and non-independent and non-identically distributed (non-IID), We examine how data variations affect model performance. 
Both dataset partitioning methods follow the guidelines established in \cite{9933818}.
\begin{figure}[htbp]
\centering \includegraphics[width=0.42\textwidth]{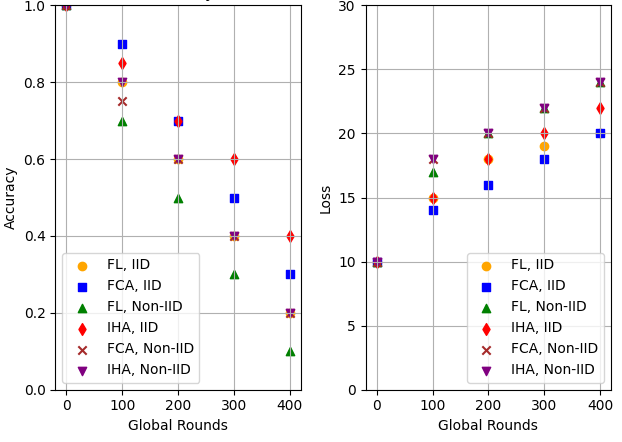}
\caption{Performance of (a) accuracy and (b) loss CIFAR-10 dataset.}
\label{CC}
\end{figure}
We utilize a straightforward convolutional neural network (CNN) model for image classification on the MNIST dataset. The model commences with a convolutional layer comprising 10 filters of size 5×5, adhering to a ReLU activation function, and 2×2 max pooling. Subsequently, the second layer of convergence incorporates 20 filters of size 5×5, along with a ReLU activation function and 2×2 max pooling. The model is further comprised of two fully connected layers with ReLU activations, resulting in ten output features.
A similar model architecture was also developed for the CIFAR-10 dataset.

\subsection{Model Efficiency}
We employ a total of 200 UAVs, where each training UAV runs mini-batch SGD once per training round. The learning rates for the MNIST and CIFAR-10 datasets are 0.035 and 0.01, respectively, while the batch size is 10. We monitor the training process until convergence, then report the results.

Fig. \ref{CC} shows the performance results for CIFAR-10. Training in CIFAR-10 poses greater challenges than MNIST, which better highlights differences between schemes. The findings show conventional FL approaches like scalable FL perform best when data is IID. However, with non-IID data and FCA, the accuracy decreases slightly. Notably, the $k$-hop scheme sees considerably slower convergence with non-IID data. This behavior can be attributed to the disparate distribution of data instances across the UAV network. This impedes the diffusion of model updates between UAVs that have distinct local datasets.

\begin{figure}[htbp]
\centering
    \includegraphics[width=0.45\textwidth]{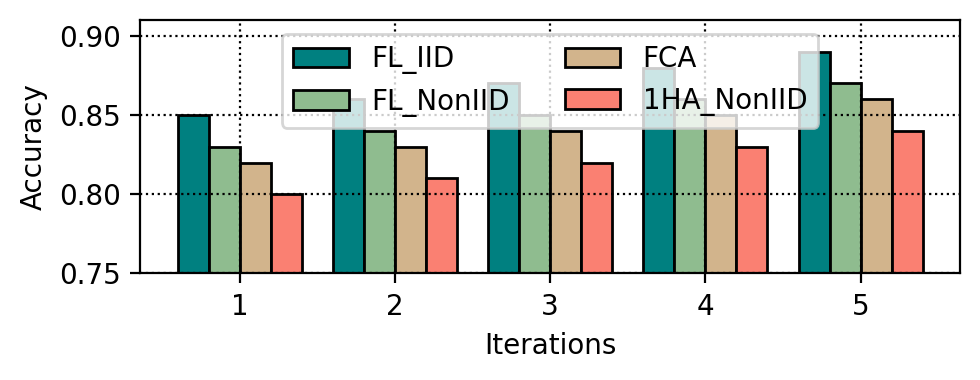}
\caption{The BCS-FL Performance of accuracy.}
\label{loss}
\end{figure}

\begin{figure}[htbp]
\centering
\includegraphics[width=0.48\textwidth]{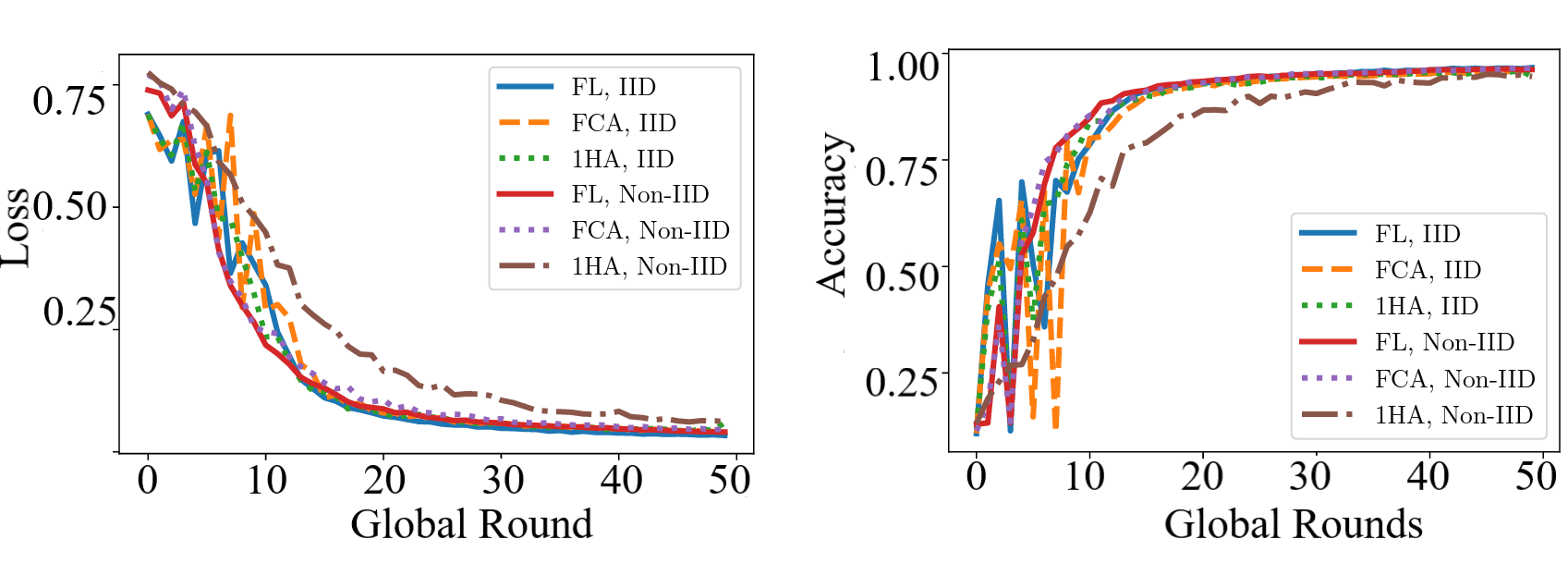}
\caption{Performance on BCS-FL (a) accuracy and (b) loss MNIST dataset.}
\label{AB}
\end{figure}

We simulate scenarios comparing suggested aggregation schemes, specifically setting $k$ = 1 for the $k$-hop. Figs. \ref{loss}-\ref{AB} show model performance metrics for MNIST using FCA and 1HA. FCA demonstrates superior performance in both IID and non-IID settings, as all UAVs participate in each global round for both schemes. However, 1HA requires approximately 50 additional rounds to match FCA's accuracy, especially for non-IID data, since FCA interconnects all UAV swarms.

\begin{figure}[htbp]
\centering \includegraphics[width=0.45\textwidth]{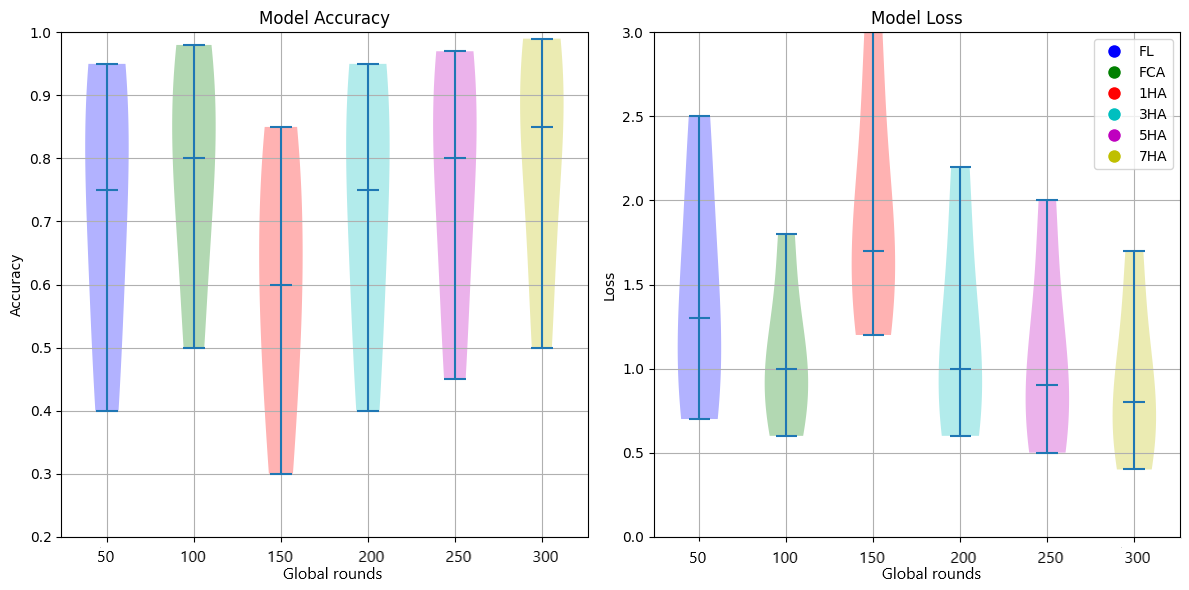}
\caption{ Influence of $k$ on BCS-FL (a) accuracy and (b) loss.}
\label{AC}
\end{figure}

Fig. \ref{AC} reveals that selecting $k$=1 results in the slowest convergence as the training information diffuses gradually. With $k$=3, performance becomes comparable to FCA. However, larger $k$ values can impose excessive communication overhead. We recommend empirically determining $k$ based on the network architecture and desired convergence/overhead trade-off.

\subsection{Communication Overhead}
\begin{figure}[htbp]
\centering \includegraphics[width=0.35\textwidth]{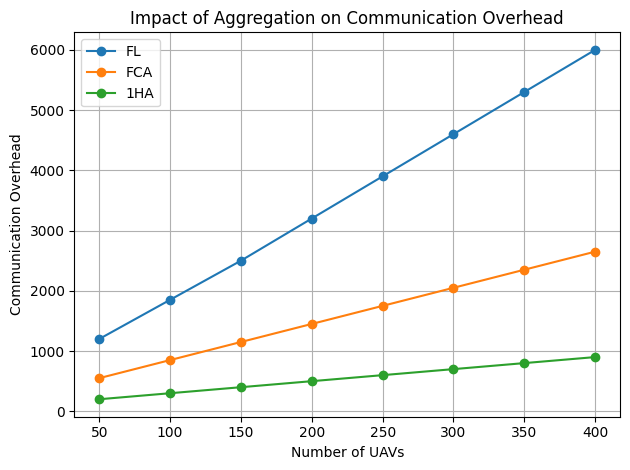}
\caption{Impact of Inter-Cluster Aggregation Scheme on Communication Overhead.}
\label{last}
\end{figure}

We conducted simulations to evaluate the communication overhead of different aggregation schemes by varying the number of UAVs ($U$). The overhead was measured by counting model update exchanges within and between clusters per training round across 20 random network layouts.
For conventional FL, the aggregation UAVs were randomly chosen. Shortest-path routing was used where possible. The results in Fig. \ref{last} show that 1HA is emerging as the most communication efficient among techniques. Despite comparable model performance to conventional FL, FCA had higher overhead. With 400 UAVs, conventional FL needed over 3,500 message exchanges, versus only ~35\% for FCA. In summary, simulations demonstrate 1HA's advantage in low overhead, offering efficient decentralized learning for UAV networks with limited resources. This provides guidance on optimizing communication efficiency versus learning performance when designing aggregation protocols.
\section{Conclusions and Future work}

 In this study, we present a novel approach tailored to UAV swarms, integrating an iterative clustering algorithm to enable effective local model aggregation and robust connectivity among CH UAVs. The hybrid iterative clustering approach groups UAVs to reduce communication overhead for model aggregation. The inter-cluster aggregation schemes of FCA and $k$-hop further minimize overhead compared to conventional FL. Simulations demonstrate BCS-FL's efficacy in learning performance, with FCA attaining accuracy comparable to centralized methods. The study provides guidance on optimizing the tradeoff between convergence and efficiency by selecting appropriate aggregation protocols. Overall, BCS-FL shows promise for collaborative learning in UAV swarms.

However, real-world deployment poses challenges, including limited onboard computing, unreliable connections between mobile UAVs, precise dynamic clustering, and coordination overhead. While discussing these limitations provides a useful perspective, our results emphasize BCS-FL innovations in scalable, decentralized FL for UAV networks. Ongoing research aims to advance BCS-FL's incentives, security, optimization, adaptability, and applicability.

\vspace{12pt}
\end{document}